\documentclass[runningheads]{llncs}

\usepackage{eccv}

\usepackage{eccvabbrv}

\usepackage{graphicx}
\usepackage{booktabs}
\usepackage{gensymb}
\usepackage{amsmath}
\usepackage{amssymb}
\usepackage{multirow}

\usepackage[accsupp]{axessibility}  

\usepackage{hyperref}

\usepackage{orcidlink}

\begin{document}

\title{CanonicalFusion: Generating Drivable \\ 3D Human Avatars from Multiple Images} 

\titlerunning{CanonicalFusion}

\author{Jisu Shin\inst{1}\orcidlink{0009-0007-7640-1512} \and
Junmyeong Lee\inst{1}\orcidlink{0009-0000-6667-1761} \thanks{Currently at POSTECH}\and
Seongmin Lee\inst{1}\orcidlink{0009-0000-8477-3286} \and
Min-Gyu Park\inst{2, 3}\orcidlink{0000-0003-1752-150X} \and \\
Ju-Mi Kang\inst{2}\orcidlink{0000-0002-0873-8558} \and
Ju Hong Yoon\inst{2, 3}\orcidlink{0000-0003-2945-8376} \and
Hae-Gon Jeon\inst{1}\orcidlink{0000-0003-1105-1666} \thanks{Corresponding author}}

\authorrunning{Shin et al.}

\institute{GIST AI Graduate School\\
\and
Korea Electronics Technology Institute (KETI)\\
\and
polygom
}

\maketitle

\begin{abstract}
We present a novel framework for reconstructing animatable human avatars from multiple images, termed CanonicalFusion. Our central concept involves integrating individual reconstruction results into the canonical space. To be specific, we first predict Linear Blend Skinning (LBS) weight maps and depth maps using a shared-encoder-dual-decoder network, enabling direct canonicalization of the 3D mesh from the predicted depth maps. Here, instead of predicting high-dimensional skinning weights, we infer compressed skinning weights, i.e., 3-dimensional vector, with the aid of pre-trained MLP networks. We also introduce a forward skinning-based differentiable rendering scheme to merge the reconstructed results from multiple images. This scheme refines the initial mesh by reposing the canonical mesh via the forward skinning and by minimizing photometric and geometric errors between the rendered and the predicted results. Our optimization scheme considers the position and color of vertices as well as the joint angles for each image, thereby mitigating the negative effects of pose errors. We conduct extensive experiments to demonstrate the effectiveness of our method and compare our CanonicalFusion with state-of-the-art methods. Our source codes are available at \url{https://github.com/jsshin98/CanonicalFusion}.
\end{abstract}

\begin{figure*}[t]
\centering
\includegraphics[width=\textwidth]{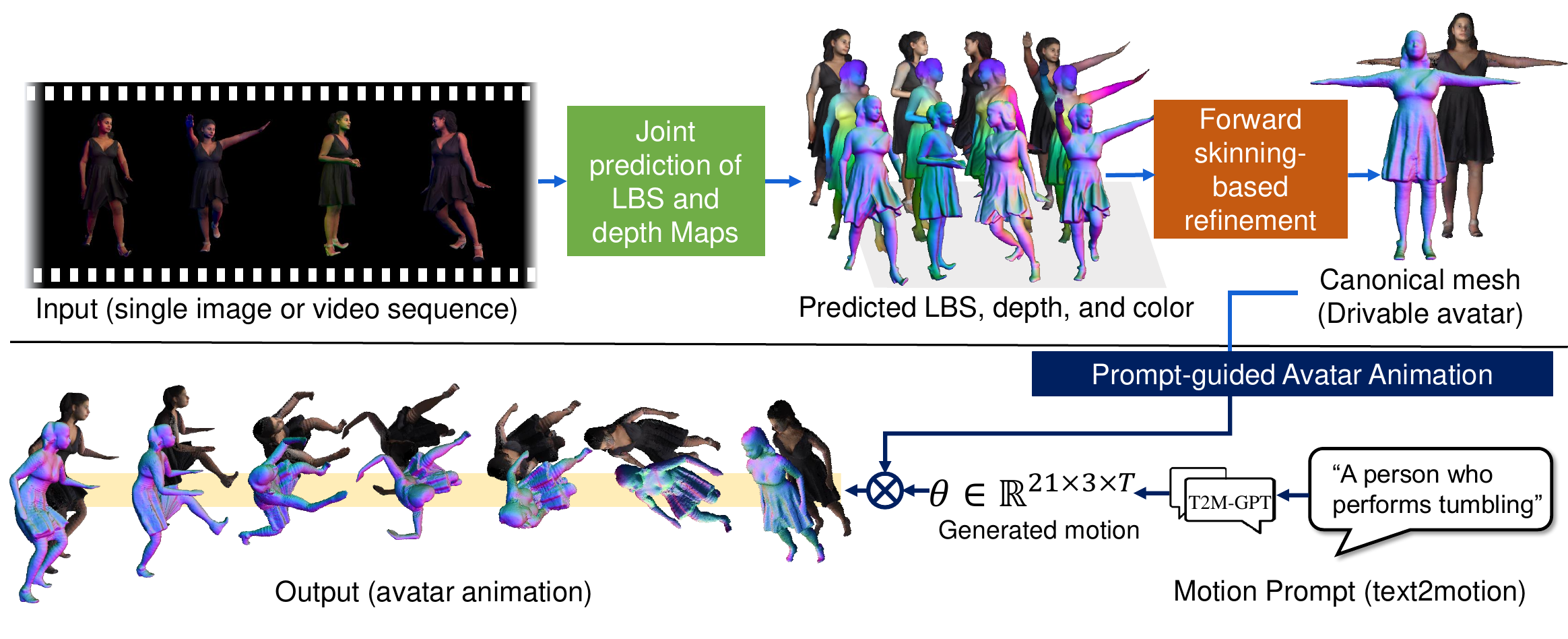}
\vspace{-7mm}
\caption{Our framework, CanonicalFusion, generates a drivable avatar from multiple images.}
\vspace{-2mm}
\label{fig:teaser}
\end{figure*}

\section{Introduction}
\label{sec:intro}
Generating human avatars from images has gained significant interest, which is one of key technologies for various applications to metaverse and AR/VR. Traditionally, this process required huge manual efforts of skilled artists with expensive equipments in controlled environments. The advances in neural networks have simplified this process, anticipating automatic avatar creation shortly.

Existing approaches address the human reconstruction problem through various methods, including training deep implicit functions~\cite{xiu2022icon, shao2022doublefield, cao2023sesdf, alldieck2022photorealistic, wang2023complete}, explicitly predicting depth maps~\cite{xiu2023econ, han2023high, smith2019facsimile}, or leveraging hybrid representations~\cite{huang2023tech} such as deformable tetrahedron~\cite{shen2021deep}. Deep implicit functions compute signed distances within a pre-defined volume where the iso-surface can be extracted from the signed distance volume or the novel view can be rendered via volume rendering. 
The explicit approaches infer explicit geometric information, such as depth maps or surface normal, allowing for higher resolution handling than the implicit approach. In addition, the reconstructed mesh or point cloud can be refined using predicted results, notably through, differentiable rendering.  
DMTet~\cite{shen2021deep} and flexicube~\cite{shen2023flexible} leverage both explicit and implicit representations as hybrid approaches. They apply both volumetric and differentiable rendering to enhance the quality of 3D models. Thanks to the huge progress, researchers have moved their eyes onto animating reconstruction as in the real world.

To bring life into a virtual human, it is essential to estimate the position of joints and skinning weights for each vertex as well as the shape and color of a human model. The majority of studies employ the skinned template models~\cite{loper2015smpl, pavlakos2019expressive} to take their skinning weights and joints as weak supervision signals.
For example, one can fit a template human model to a reconstructed mesh and assign skinning weights to each vertex using nearest neighbor search. We refer to this process as inverse skinning that assigns skinning weights to the posed mesh, allowing for the acquisition of a canonical mesh by warping each vertex to canonical space. This can serve as a post-processing step for animating human mesh, and the weights can be further refined by using multiple scan models as in SCANimate~\cite{saito2021scanimate}. Moreover, several studies~\cite{weng2019photo, yang2021s3} attempt to directly predict skinning weights in the image space together with depth values. The predicted depth map back-projects pixels into 3D points, and these points are animated using the predicted skinning weights.
Here, the dimension of skinning weights aligns with the number of joints and typically exhibits sparsity (\ie, zero values) as only a few joints influence the position of one vertex. 

On the other hand, numerous studies have evolved the canonical template mesh while changing the posture, using a single image~\cite{hu2023sherf, he2021arch++, huang2020arch, dong2022pina, huang2023one, liao2023high}, multiple frames~\cite{peng2024animatable, jiang2022selfrecon, zheng2022structured, yu2023monohuman, guo2023vid2avatar, iqbal2023rana, zielonka2023drivable}, or generated images~\cite{kim2023chupa} through diffusion networks. This approach has double-sided edges, which retains the robustness of reconstruction, but often fails to recover loose clothing and hair. 

In this work, we focus on generating drivable 3D human avatars from multiple images by jointly predicting depth and Linear Blend Skinning (LBS) maps, followed by a forward skinning-based differentiable rendering. 
Similar to sandwich-like approaches~\cite{xiu2023econ, gabeur2019moulding, smith2019facsimile}, we initially infer double-sided depth and LBS maps for each image. Here, we utilize a stacked autoencoder MLP model to represent the LBS weights as a low-dimensional vector and decode them back to full-dimensional skinning weights, \ie, 55 dimensions. We can reconstruct the canonical mesh directly from an input image, predicted depth map, and predicted LBS map.
After generating the initial canonical mesh, we refine the position and color of vertices through forward skinning-based differentiable rendering. Our method warps the canonical mesh \wrt multiple images via forward skinning and renders the normal maps and color images using a differentiable rasterizer. Then, we minimize the color and normal map errors to enhance the geometrical accuracy of the mesh; moreover, we consider the error of the joints, \ie, 3D human pose parameters, as one of the loss functions, which comes from the inconsistency between the initial pose and the reconstructed mesh. 

Our pipeline, dubbed \textit{CanonicalFusion}, can animate by using motions generated from texts~\cite{zhang2023t2m}, as shown in \cref{fig:teaser}. Furthermore, we conducted an extensive ablation study to validate each component of our method with publicly available benchmarks, demonstrating superior performance compared to the previous baseline. We also show the in-the-wild results to validate the practical utility.

\section{Related Works}
We review relevant studies by categorizing them into two groups: clothed human reconstruction and drivable avatar generation from images.

\subsection{Clothed Human Reconstruction}
Existing 3D clothed human reconstruction methods can be roughly divided into two major categories depending on their 3D representation: implicit and explicit. Implicit approaches train a deep implicit function defined for a continuous 3D space, that is capable of representing arbitrary humans. PIFu and PIFuHD~\cite{saito2019pifu, saito2020pifuhd} are pioneering studies that exploit pixel-aligned image features to predict the occupancy value and color at a queried position. Following studies~\cite{he2020geo, he2021arch++, huang2020arch, xiu2022icon, zheng2021pamir, dong2022pina, yang2021s3, li2020robust, liao2023high} incorporate geometric priors such as depth, LiDAR, 3D volumetric features, or parametric body models to alleviate the limitations of PIFu. For example, Geo-PIFu~\cite{he2020geo} employs latent voxel features to generate occupancy volume coarsely. ARCH~\cite{huang2020arch} and ARCH++~\cite{he2021arch++} utilize the geometry encoder to estimate the 3D body shape of a given subject. 

Explicit methods predict explicit geometric representation such as depth maps~\cite{gabeur2019moulding} or point clouds~\cite{ma2021power} to reconstruct 3D humans. Predicting front and back depth maps of given RGB images and then merging them in a "sandwich-like" manner~\cite{gabeur2019moulding} to form explicit meshes became a popular approach because of its efficiency in predicting 3D human models. To further generate high fidelity 3D clothed humans, several studies~\cite{smith2019facsimile, xiu2023econ} derive normal maps from depth maps facilitating the reconstruction of loose clothes and challenging postures. Recent approaches incorporate diffusion with text-level guidance~\cite{huang2023tech, kim2023chupa} or latent code inversion from given reference image~\cite{xiong2023get3dhuman}. With the generated normal maps with text prompts, differentiable rendering~\cite{worchel2022multi} is used to optimize the clothed human mesh from skinned body models~\cite{loper2015smpl, pavlakos2019expressive}.

Recently, on the other hand, the boundary between implicit and explicit approaches has disappeared thanks to the hybrid representations~\cite{shen2021deep, shen2023flexible}. The hybrid representations provide differentiable mesh generation with optimized implicit volumes, therefore, both volume rendering and differentiable rendering can be applicable. TeCH~\cite{huang2023tech}, for example, exploits deep marching tetradedron~\cite{shen2021deep} to reconstruct human models. 

\subsection{Drivable Human Avatar Generation} 
Drivable 3D human avatars can be generated from pre-scanned models~\cite{saito2021scanimate}, images~\cite{he2021arch++, huang2020arch, huang2023one, liao2023high}, and texts~\cite{hong2022avatarclip}, where we focus on generating them from images including a single image input.
Several studies~\cite{he2021arch++, huang2020arch, huang2023one, liao2023high} train animatable implicit neural representations from a single image with the guidance of the SMPL body model.
ARCH~\cite{huang2020arch} and ARCH++~\cite{he2021arch++} assign skinning weights to the canonical surface from the implicit skinning field, initialized with skinning weights from the underlying canonical SMPL model. 
However, distant points, \eg, loose cloth, from the surface of a canonical template model are not correctly initialized, resulting in poor results. CAR~\cite{liao2023high} addresses this by using front and back normal maps as geometric cues. They learn the canonical implicit representation by adopting a canonical Signed-Distance Field (SDF) and further refines normal maps to generate high-fidelity clothed avatars.

Another group of studies takes an explicit approach that directly estimates the skinning weight map from a given image.
Weng \etal \cite{weng2019photo} propose to learn an explicit warping function that warps normal and skinning maps of a predicted SMPL model into the silhouette of an image and S3~\cite{yang2021s3} predicts the skinning field along with occupancy and pose fields, enabling reconstruction and animation of clothed meshes simultaneously.

Recent methods take multi-view images~\cite{peng2024animatable}, videos~\cite{jiang2022selfrecon, zheng2022structured, yu2023monohuman, guo2023vid2avatar, zielonka2023drivable, dong2022pina}, or multiple scans~\cite{saito2021scanimate, shen2023x, li2022avatarcap, chen2021snarf} as inputs to enhance the quality of drivable avatar. For example, SCANimate~\cite{saito2021scanimate} ensures cycle consistency between posed scans and canonical scans by utilizing an implicit skinning field to transform posed scans into canonical scans and back under the guidance of the body model. 
SNARF\cite{chen2021snarf} proposed a differentiable forward skinning field representation, which is trained without a canonical avatar using an iterative root-finding algorithm. However, they require detailed 4D scanned data such as CAPE\cite{CAPE:CVPR:20} to train a forward skinning network.

\section{The Proposed Method}
\label{sec:proposed}
\begin{figure*}[t]
\centering
\includegraphics[width=\textwidth]{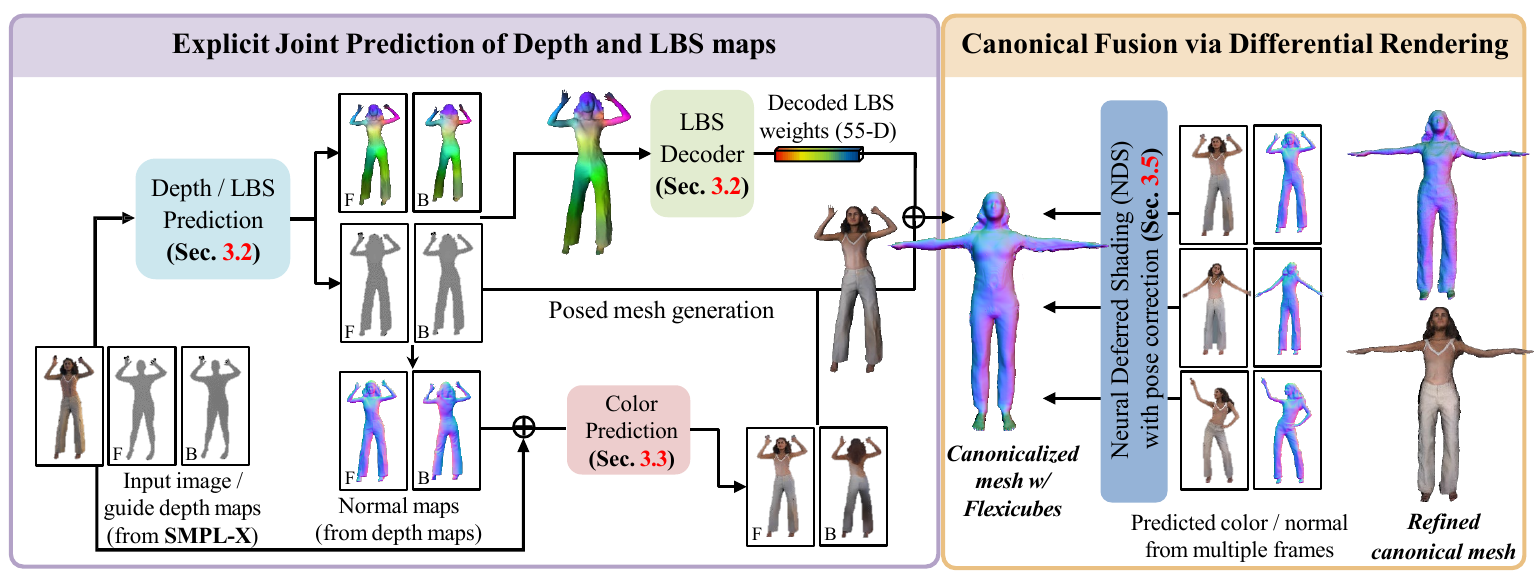}
\vspace{-7mm}
\caption{An overview of our framework, CanonicalFusion. It takes RGB image and depth maps generated from SMPL-X and estimates dual-sided depth and 3-dimensional LBS weight map. Original skinning weights are decoded from compressed LBS weight maps and used to generate a canonicalized mesh. To further increase the quality, canonical mesh is refined by integrating multiple frames with forward skinning based differentiable rendering.}
\label{fig:overview}
\end{figure*}

Our pipeline consists of two main steps in \cref{fig:overview}. First, we jointly predict geometry and skinning weights from a single input image followed by an additional texture prediction network. Afterward, we generate an initial mesh from the predicted results and then canonicalize the mesh to feed the canonical mesh to the next step. The second step involves the forward skinning-based differentiable rendering, refining the canonical mesh while minimizing geometric and photometric errors between the predicted and the rendered images. Note that our framework, CanonicalFusion, has no restrictions on the number of images, viewpoints, and pose variations. 

\subsection{Preliminaries on Linear Blend skinning}
\label{subsec:preliminary}
Since the template model, \ie, SMPL-X~\cite{pavlakos2019expressive}, can encompass various characteristics of a human avatar such as shape, pose, joints, semantic body parts, and skinning weights, we leverage its concept to define skinning weights for clothed humans.
The position of vertices changes \wrt $\theta$ and $\beta$ in which $\beta$ is an N-dimensional vector representing the shape and $\theta \in \mathbb{R}^{J\times 3}$ refers to pose parameters, \ie angles for each joint, with $J$ denoting the number of joints, \ie 55. $\theta$ can be converted to a set of rotation matrices, $\mathbf{T}=\{T_i \in \mathbb{R}^{3\times3}, i=1,...,J\}$. Given a mesh $\mathcal{M} = \{\mathcal{V}, \mathcal{F}\}$, where $\mathcal{V}$, $\mathcal{F}$ denotes vertices and faces, respectively, the position of joints $\mathbf{J} = \{ J_i \in \mathbb{R}^{3}, i=1,...,J\}$ can also be determined by regressing designated vertices in $\mathcal{V}$. Here, $i$ of $T_i$ and $J_i$ indicates an index of a joint. Once pose parameters are determined, $\mathcal{V}$ can be moved accordingly,
\begin{equation}\label{eq:lbs_equation}
\begin{split}
    \mathbf{x}^p &= \text{LBS}(\mathbf{x}^c, {\mathbf{T}})=\left( \sum_{i}{w_i(\mathbf{x})T_{i}} \right) \mathbf{x}^c,  \\
    \mathbf{x}^c &= \text{LBS}^{-1}(\mathbf{x}^p, {\mathbf{T}})=\left( \sum_i {w_i(\mathbf{x})T_{i}} \right)^{-1} \mathbf{x}^p
\end{split}
\end{equation}
where $\text{LBS}(\cdot)$ warps a point as a linear combination of transformed vertices by $\mathbf{T}$ and $\mathbf{x}^c$, $\mathbf{x}^p$ are vertices in canonical and posed space respectively.
The inverse of $\text{LBS}(\cdot)$ transforms a point from the posed to the canonical space. $w_i(\mathbf{x})$ is a linear blend skinning (LBS) weight of vertex $\mathbf{x}$ for the $i^{\text{th}}$ joint, and $\mathbf{x}^c$, $\mathbf{x}^p$ are vertex in canonical and posed space respectively.

\begin{figure}[t]
\centering
\includegraphics[width=1.0\columnwidth]{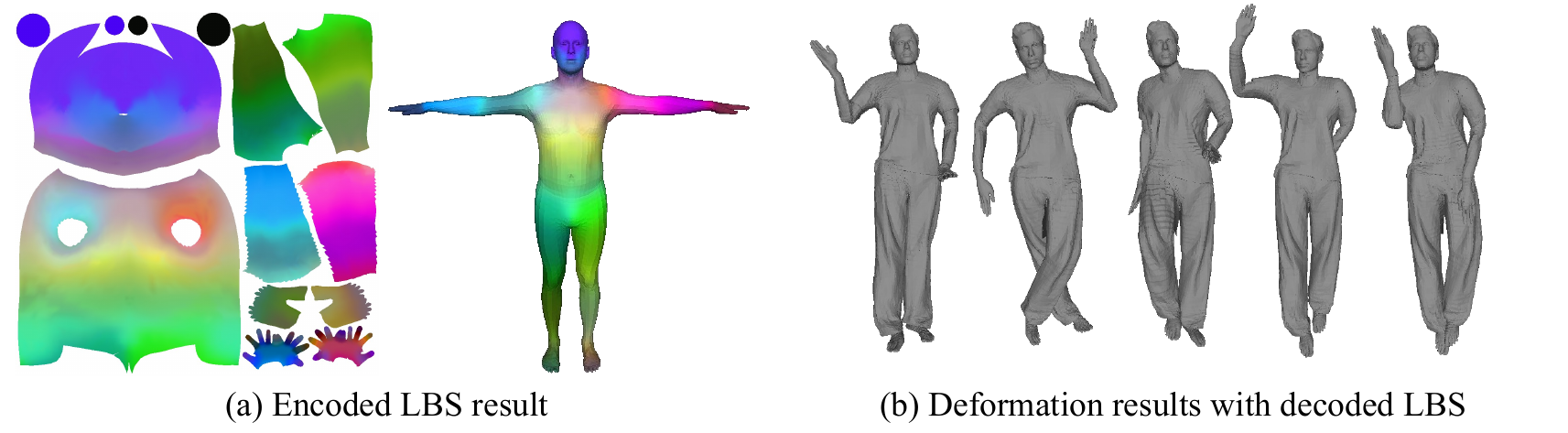}
\vspace{-8mm}
\caption{(a) UV map and SMPL mesh colored with encoded skinning weights. (b) Reposed mesh using decoded LBS weight from the pretrained decoder.}
\label{fig:lbs}
\end{figure}

\subsection{Joint Depth and LBS Prediction}
Given an RGB image, we first predict the depth and skinning weight maps for both the front and back side views using a shared-encoder-double-decoder network, denoted as $\mathcal{S}(\cdot)$, employing the ATUNet architecture~\cite{oktay2018attention}, 
\begin{equation}
(\mathbf{\hat{D}}, \mathbf{\hat{L}}) = \mathcal{S}(\mathbf{I}, \mathbf{\bar{D}})
\end{equation}
where $\mathbf{I}$ is an input image and $\mathbf{\bar{D}}$ is dual-sided depth maps generated from the predicted SMPL-X parameters. Here, the network $\mathcal{S}(\cdot)$ has a shared encoder and two decoder networks in which the first decoder predicts depth maps $\mathbf{\hat{D}}$ and the second one infers LBS maps $\mathbf{\hat{L}}$. Due to the sparsity nature of skinning weights, we embed these skinning weights into a low-dimensional space to effectively compute linear blend skinning (LBS) weights. Then, with the aid of LBS weights and depth maps, it becomes feasible to canonicalize the predicted mesh directly. Lastly, we estimate color images using an additional network, which takes input images and dual-sided normal maps as input which are converted from the predicted depth map.

\label{subsec:jointprediction}
\subsubsection{Compact representation of skinning weights.}
\label{subsubsec:lbs}
We employ a stacked autoencoder MLP network to compress skinning weights into a 3-dimensional latent space considering the sparsity nature of skinning weights. Here, we denote the low-dimensional LBS weight map as $\mathbf{L}$ and the full-dimensional LBS weight map as $\mathbf{L}^F$.
We utilize the SMPL-X model's skinning weights, which are 55-dimensional, and interpolate skinning weights in the SMPL-X's UV coordinate via barycentric interpolation, where the resolution of the UV map is $1024\times1024$. Consequently, the number of training samples to train the autoencoder network is about 800K. With these training samples, we minimize the following objective function:
\begin{equation}\label{eq:decoder}
\mathcal{L}_\mathcal{P} = \lambda_{\mathrm{L1}}L_1(\mathbf{L}^F, \mathbf{\hat{L}}^F) + \lambda_{\mathrm{NZ}}L_{\mathrm{nonzero}}(\mathbf{L}^F, \mathbf{\hat{L}}^F) + \lambda_{\mathrm{KL}}L_{KL}(\mathbf{L}^F, \mathbf{\hat{L}}^F),
\end{equation}
where $L_{\mathrm{nonzero}}$ ensures that the estimated LBS weight map, $\mathbf{\hat{L}}^F$, has the same number of nonzero elements as the ground truth weights, $\mathbf{L}^F$, after decoding back to the original dimension. We use the approximation for the nonzero count using differentiable relaxation by employing a sum of narrow Gaussian basis functions. $L_1$ represents the $L_1$ loss and $L_{KL}$ is the KL divergence loss. Hyperparameters $\lambda_{\mathrm{L1}}$, $\lambda_{\mathrm{NZ}}$ and $\lambda_{\mathrm{KL}}$ are empirically set to 1.0, 0.1 and 0.1, respectively. Since the sum of skinning weights is 1, we apply the softmax operation after the last layer of MLP. \cref{fig:lbs}(b) shows that our skinning weight can deform the clothed mesh naturally, which is decoded back from the latent space. \cref{fig:lbs}(a) provides a visualization of the latent code.
Note that this LBS compression network is pre-trained before training $\mathcal{S}(\cdot)$. During inference, we only utilize the decoder to decode the predicted LBS weights.

\subsubsection{Objective Function.}
Our loss function for the initial mesh prediction network $\mathcal{S}(\cdot)$ is defined as:
\begin{equation}\label{eq:overall}
\begin{split}
\mathcal{L}_{\mathcal{S}} =& \lambda_1 L_{L2}(\mathbf{D}, \mathbf{\hat{D}}) +  \lambda_2 L_{\text{ssim}}(\mathbf{D}, \mathbf{\hat{D}})
+ \lambda_3 L_{cos}(\mathbf{N}, \mathbf{\hat{N}})  \\
& + \lambda_4 L_{L2}(\mathbf{N}, \mathbf{\hat{N}}) + \lambda_5 L_{L2}(\mathbf{L}, \mathbf{\hat{L}}),
\end{split}
\end{equation}
where $\mathbf{D}$ is the ground truth depth map and $\mathbf{N}$ is the normal map calculated from the depth map. Similarly, $\mathbf{L}$ denotes the compressed LBS weight map. We utilize $L_2$, structural dissimilarity, and cosine loss functions to compare depth, normal, and LBS weight maps. Note depth maps and LBS weight maps are predicted for the front and hidden views. In addition, $\lambda_1$, $\lambda_2$, $\lambda_2$, $\lambda_3$, $\lambda_4$ and $\lambda_5$ are set to $0.9$, $0.1$, $1$, $0.15$, and $1$, respectively.

\subsection{Texture Prediction}

Given the pretrained network $\mathcal{S}(\cdot)$, we train the color prediction network $\mathcal{C}(\cdot)$ to infer shade-removed images for the front and hidden surfaces. We employ a UNet architecture that takes an input front image and the normal maps from the predicted depth map, 
\begin{equation}
\mathbf{\hat{C}} = \mathcal{C}(\mathbf{I}\oplus \mathbf{\hat{N}}), 
\end{equation}
where $\oplus$ indicates a channel-wise concatenation and $\mathbf{\hat{N}}$ is the predicted normal map. 
To train $\mathcal{C}(\cdot)$, we minimize 
\begin{equation}
\mathcal{L}_{\mathcal{C}} = \lambda_{L2} L_{L2}(\mathbf{C}, \mathbf{\hat{C}}) + \lambda_{\text{VGG}} L_{\text{VGG}}(\mathbf{C}, \mathbf{\hat{C}}),
\end{equation}
where $\mathbf{\hat{C}}$ and $\mathbf{C}$ are the predicted and ground truth color images for front and back views, respectively.
$\lambda_{L2}$ and $\lambda_{\text{VGG}}$ are hyperparameters for the $L_2$ loss and the perceptual loss which are set to $0.85$ and $0.15$, respectively.

\subsection{Canonical Mesh Reconstruction} 
The reconstructed avatar from $\mathbf{\hat{D}}$ is defined in the posed space. With the predicted LBS maps $\mathbf{\hat{L}}$, it can be warped to its canonical space without an additional optimization procedure to fit the template human model.
Here, we define the canonical mesh as $\mathcal{M}^c = \{\mathcal{V}^c, \mathcal{C}, \mathcal{F} \}$, consisting of a set of vertices, vertex colors, and faces, which is obtained via an inverse skinning operation from the posed mesh, $\mathcal{M}^p = \{\mathcal{V}^p, \mathcal{C}, \mathcal{F} \}$,
\begin{equation}
\mathcal{V}^c = \{{\text{LBS}}^{-1}(\mathbf{x}^p, \mathbf{T}) \text{ } | \text{ } \forall \mathbf{x}^p\ \in \mathcal{V}^p \},
\end{equation}
where ${\text{LBS}}^{-1}(\cdot)$ utilizes the skinning weights decoded back from the encoded LBS map, $\mathbf{\hat{L}}$. However, this inverse skinning process fails to recover unseen geometry, such as armpit and thigh regions where the reconstructed mesh is not separated. Therefore, if we remove the faces stretched above the pre-defined value, these regions remain empty in the canonical space. 
To fill these empty regions, we combine the canonicalized mesh and canonical template mesh through signed distance integration as below: 
\begin{equation}\label{eq:sdf_functions}
\begin{split}
\mathbf{S}_c(\mathbf{x}^c) &= {SDF}(\mathbf{x}^c, \partial\mathcal{M}^c_\text{SMPL-X}), \\
\mathbf{S}_p(\mathbf{x}^c) &= {SDF}({LBS}(\mathbf{x}^c, \mathbf{T}), \partial\mathcal{M}^p) = {SDF}(\mathbf{x}^p, \partial\mathcal{M}^p), 
\end{split}
\end{equation}
where $SDF(\cdot)$ computes the signed distance from 3D point $\mathbf{x}^c$ to the nearest surface of $\partial\mathcal{M}$, $\partial\mathcal{M}$ represents the surface of mesh $\mathcal{M}$, and $\mathcal{M}^c_\text{SMPL-X}$ is the canonical template mesh. To integrate two signed distance functions in the canonical space, we integrate signed distance values within a pre-defined volume, \ie, $2m^3$, where the LBS weights are obtained by taking the LBS weights of the nearest vertex of $\mathcal{M}^c_\text{SMPL-X}$. Then, we warp $\mathbf{x}^\mathbf{c}$ to the posed space with the LBS weights and compute the signed distance in the posed space. Here, direct canonicalization of $\mathcal{M}^p$ generates holes and artifacts, therefore, we warp $\mathcal{M}^c$ back to the posed space to sample the signed distance values. Finally, we take the signed distance of $\mathbf{S}_p(\mathbf{x}^c)$ if there exists canonical vertices $\mathcal{V}^c$ in the vicinity of $\mathbf{x}^c$,
\begin{equation}
\label{eqs:sdf_integration}
    \begin{array}{l}
     \mathbf{S}(\mathbf{x}^c) = 
\left\{
\begin{array}{cc}
    \mathbf{S}_p(\mathbf{x}^c) & \quad \text{min}|\mathbf{x}^c-\mathcal{V}^c| < \tau, \\ 
    \mathbf{S}_c(\mathbf{x}^c) & \quad \text{otherwise}, \\
\end{array}
\right.
    \end{array}
\end{equation}
where $\tau$ is empirically set to $5$ by observing average discrepancy between ground truth SMPL-X and posed meshes.

Afterward, we obtain a canonical mesh via the marching cubes and re-parametrize it using Flexicubes~\cite{shen2023flexible}, transforming it into a differentiable, watertight and compact mesh via a gradient-based optimization. Here, a reference frame with the smallest Chamfer distance over the canonical SMPL-X mesh for multiple input images is chosen as the initial canonical mesh.
In the rest of the paper, we refer $\mathcal{M}^c$ as the canonical mesh after applying Flexicubes.

\subsection{Incorporating Multiple Images into \\ Forward Skinning-based Differentiable Rendering} 
\label{subsec:diff_rend}
As the final step, we rasterize the posed mesh deformed from the initial canonical mesh using Neural Deferred Shading (NDS)~\cite{worchel2022multi} to minimize the discrepancy between the rendered image and the predicted image from the previous step. In this process, we refine the 3D human pose as well as the shape and color of the canonical mesh. This is crucial because errors in human pose can substantially impair the quality of deformation results generating undesirable artifacts.

Formally, we have depth maps, $\mathbf{D}=\{\mathbf{D}_1, \cdots, \mathbf{D}_n\}$, color images, $\mathbf{C} = \{\mathbf{C}_1, \cdots, \mathbf{C}_n\}$, posed meshes, $\mathcal{M}^P = \{\mathcal{M}^p_1, \cdots, \mathcal{M}^p_n\}$ and canonical mesh $\mathcal{M}^c$ obtained from the previous steps, where $\mathbf{D}$ and $\mathbf{C}$ are double-sided depth maps and images and $n$ is the number of images. Here, the vertex color and faces of the canonical mesh $\mathcal{M}^c$ (selected in \cref{subsec:jointprediction}) are invariant \wrt the change of vertex positions. With the linear skinning weights of the canonical mesh, we can apply forward skinning-based warping w.r.t given transformation as follows:
\begin{equation}
\mathcal{\hat{V}}_i^p = \{{\text{LBS}}(\textbf{x}^c, T_i) \text{ } | \text{ } \forall \mathbf{x}^c\ \in \mathcal{V}^c \}, 
\end{equation}
where $\mathcal{\hat{M}}_i^p = \{\mathcal{V}_i^p, \mathcal{C}, \mathcal{F} \}$ is the predicted posed mesh for $i^{\text{th}}$ image via forward warping and $T_i$ is the transformation matrix calculated from the 3D human pose. 
Before the optimization, we downsample the canonical mesh and progressively upsample the mesh while updating the canonical mesh. 

To this end, we define our objective function,  
\begin{equation}
\label{eq:diff_render}
\begin{split}
\mathcal{L} =& \lambda_{1} L_{\text{laplacian}}(\mathcal{M}^c) + \lambda_{2} L^{\text{reg}}_{\text{normal}}(\mathcal{M}^c) \\
&+ \sum_{i=1}^{n} \big( \lambda_{3}L_{\text{normal}}(\mathbf{N}^p_i, \mathbf{\hat{N}}^p_i) + \lambda_{4} L_{\text{mask}}(\mathbf{M}^p_i, \mathbf{\hat{M}}^p_i),  \\
&\qquad +\lambda_{5}L_{\text{chamfer}}(\mathcal{M}^p_i, \mathcal{\hat{M}}^p_i) + \lambda_{6} L_{\text{color}}(\mathbf{C}^p_i, \mathbf{\hat{C}}^p_i) 
\big), 
\end{split}
\end{equation}
which updates the canonical mesh $\mathcal{M}^c$ and 3D human pose $\mathbf{T}$ using differential renderer~\cite{laine2020modular}. 
$L_{\text{normal}}$ minimizes the $L_1$ loss between normal maps, and $L_{\text{mask}}$ and $L_{\text{color}}$ minimize the discrepancy between binary masks and color images using MSE-based losses, respectively. Note that our refinement procedure can take an arbitrary number of input images as input, which means $n$ can range from at least 1 to tens of images. 
To prevent the abrupt change of vertices, we adopt the Laplacian loss function in NDS~\cite{worchel2022multi}, 
\begin{equation}
    L_{\text{laplacian}} = \frac{1}{m} \sum^m_{i=1} \| \delta_i \|^2_2,
\end{equation}
where $\delta_i=(L\mathcal{V}^c)_i \in \mathbb{R}^3$ are the differential coordinates of vertex $i$ with the graph Laplacian $L$.
We also utilize normal consistency loss as: 
\begin{equation}
    L^{\text{reg}}_{\text{normal}}=\frac{1}{|\mathcal{F}|}\sum_{(i,j)\in\mathcal{\bar{F}}}(1-n_i \cdot n_j)^2,
\end{equation}
where $|\mathcal{F}|$ is the number of the faces, $\mathcal{F}$, which is paired with a shared edge and $n_i$ $\in$ $\mathbb{R}^3$ is the normal of triangle $i$ in $\mathcal{M}^c$. We further minimize the Chamfer distance~\cite{johnson2020accelerating} between $\mathcal{\hat{M}}^p_i$ and $\mathcal{M}^p_i$ for the $i^{\text{th}}$ frame. Due to the potential misalignment between each 3D human pose and the reconstructed mesh, we update the initial human pose parameters to consider pose errors while updating the shape and color of the canonical mesh.  

We initially set the hyperparameters $\lambda_1$, $\lambda_2$, $\lambda_3$, $\lambda_4$, $\lambda_5$, $\lambda_6$ to $40$, $0.1$, $1$, $2$, $0.0001$, and $0$ respectively. We employ a progressive update scheme in NDS~\cite{worchel2022multi} for $\mathbf{T}$ and $\mathcal{V}^c$, where we upsample $\mathcal{V}^c$ by a factor of about $4$ and multiply $\lambda_3$ and $\lambda_4$ by $4$ every $500$ iterations, up to $2000$ iterations.
After optimizing $\mathbf{T}$, we fix it and proceed to recover $\mathcal{C}$ and $\mathcal{V}^c$ by minimizing the $L_2$ loss between the input color maps, $\mathbf{C}^p_i$, and the predicted color maps, $\mathbf{\hat{C}}^p_i$, for the $i^{\text{th}}$ frame, denoted as $L_{\text{color}}$. During this step, we set $\lambda_6$ to $10$ to prioritize the color optimization.

\section{Experimental Results}
\label{sec:exp}
\noindent\textbf{Datasets. }
We employed popularly used commercial~\cite{rp} and public~\cite{yu2021function4d} datasets. First, we obtained 412 human models from RenderPeople~\footnote{https://renderpeople.com/}, denoted as RP. The ground truth SMPL parameters for RP dataset were sourced from Agora~\cite{patel2021agora}. Within 412 models, 332 models are mostly standing, whereas 80 models are in T-pose. To augment our training data, we generated 800 human models from the 80 T-posed models while changing body poses using V-poser~\cite{pavlakos2019expressive}. In addition, we used the THuman 2.0~\cite{yu2021function4d} dataset for training our model, containing 526 models. For testing, we selected 20 scan models from RP and TH3.0 datasets that were not used for training. The list of tested models is given in the supplementary material. 

\noindent\textbf{Training data generation.}
To generate training data, we followed the protocols of prior studies~\cite{alldieck2022photorealistic, saito2020pifuhd} and rendered synthetic images and depth maps as follows. Initially, human models were centered at the origin aligned based on the centroid of each mesh to ensuring them to be positioned at the center of the image. Afterward, we put a perspective camera with 52 degrees of field-of-view that is 3 meters away from the origin. Next, the human models were rotated $0$, $1$, $2$, $3$, $4$, $5$, $10$, $15$, $20$, $30$, $40$, $60$, $90$, $120$, $150$, $180$, $210$, $240$, $270$, $300$, $330$, $340$, $345$, $350$, $355$, $356$, $357$, $358$ and $359$ degrees, along the vertical axis. To synthesize images, we sampled spherical harmonics mimicking natural illuminations~\cite{saito2019pifu} and rendered 20 images with different lighting conditions. Ground truth data includes color albedo maps, depth maps, and LBS maps for both the front and back sides, and silhouette masks. The resolution of the image is $512 \times 512$. 

\noindent\textbf{Training settings.}
We used the Adam~\cite{kingma2014adam} optimizer with $\beta_1$=$0.9$ and $\beta_2$=$0.999$, and set a learning rate to $0.001$. We trained our main network (\cref{subsec:jointprediction}) using four NVIDIA RTX 3090 GPUs which took 2 days for training. Our pipeline takes about $11$ minutes in total: $3$ minutes for the initial canonicalization and $8$ minutes for the forward skinning-based differentiable rendering.

\begin{table*}[t]
\setlength{\tabcolsep}{7pt}
\centering
\caption{Quantitative comparison of monocular human reconstruction methods. $^{*}$ denotes the ground truth SMPL(-X) parameters were used as input. PIFu$^{*}$ and PaMIR$^{*}$ are newly devised in ICON~\cite{xiu2022icon} to analyze the dependency on the template model. $\dagger$ and $\ddagger$ denote our result trained on TH2.0 and combination of TH2.0 and RP, respectively. Average point-to-surface distance, chamfer distance, and average surface normal error are denoted as P2S, CF and NR, respectively. The first(dark color) and second-best(light color) results are marked in color.}
\label{table:eval_posed}
\begin{center}
\footnotesize
\vspace{-8mm}
\resizebox{\linewidth}{!}{%
	\begin{tabular}{l | c | c | c | c | c | c | c | c | c | c }
		\noalign{\hrule height 1pt}
        & \multirow{2}{*}{DATASET} & \multicolumn{3}{c|}{RP} & \multicolumn{3}{c|}{2K2K} & \multicolumn{3}{c}{TH3.0} \\ \cline{3-11}
        & & {P2S$\downarrow$} & {CF$\downarrow$} & {NR$\downarrow$} 
        & {P2S$\downarrow$} & {CF$\downarrow$} & {NR$\downarrow$} 
        & {P2S$\downarrow$} & {CF$\downarrow$} & {NR$\downarrow$} \\\hline
        PIFuHD~\cite{saito2020pifuhd} & RP
        & 1.420 & 1.434 & 1.401
        & 1.334 & 1.399 & 1.266
        & 1.534 & 1.527 & 1.628\\
        PIFu$^{*}$~\cite{saito2019pifu} & RP
        & 1.787 & 1.816 & 1.822
        & 1.390 & 1.433 & 1.764
        & 1.747 & 1.742 & 1.820\\
        PaMIR$^{*}$~\cite{zheng2021pamir} & RP
        & 1.545 & 1.616 & 1.679
        & 1.464 & 1.536 & 1.654
        & 1.706 & 1.722 & 1.715\\
        ICON$^{*}$~\cite{xiu2022icon} & RP
        & 1.296 & 1.364 & 1.447
        & 1.582 & 1.675 & 2.024
        & 1.371 & 1.437 & 1.666\\ 
        PHORHUM~\cite{alldieck2022photorealistic} & CUSTOM
        & 1.512 & 1.617 & 1.633
        & 1.638 & 1.663 & 2.206
        & 2.073 & 2.120 & 2.716\\
        ECON$^{*}$~\cite{xiu2023econ} & TH2.0
        & 2.066 & 2.290 & 1.756
        & 2.382 & 2.769 & 1.806
        & 1.944 & 2.139 & 1.821\\
        2K2K$^{*}$~\cite{han2023high} & TH2.0+RP
        & 1.097 & 1.195 & 1.507
        & 1.310 & 1.265 & 1.339
        & 1.416 & 1.542 & 1.952\\
        TeCH$^{*}$~\cite{huang2023tech} & N/A
        & 1.489 & 1.523 & 2.068
        & 1.428 & 1.467 & 1.808
        & 1.721 & 1.795 & 2.551 \\\hline
        Ours$^{*}$ & RP
        & \cellcolor{blue!10}0.901 & \cellcolor{blue!10}0.978 & \cellcolor{blue!20}0.976
        & \cellcolor{blue!10}0.914 & \cellcolor{blue!10}1.007 & \cellcolor{blue!12}0.761
        & 1.086 & 1.189 & 1.413 \\
        Ours$^{*}$$^\dagger$ & TH2.0
        & 1.043 & 1.126 & 1.049
        & 1.011 & 1.102 & 0.862
        & \cellcolor{blue!10}1.074 & \cellcolor{blue!10}1.169 & \cellcolor{blue!10}1.281 \\
        Ours$^{*}$$^\ddagger$ & TH2.0+RP
        & \cellcolor{blue!20}0.886 & \cellcolor{blue!20}0.943 & \cellcolor{blue!10}0.987
        & \cellcolor{blue!20}0.897 & \cellcolor{blue!20}0.981 & \cellcolor{blue!10}0.793
        & \cellcolor{blue!20}1.072 & \cellcolor{blue!20}1.165 & \cellcolor{blue!10}\cellcolor{blue!20}1.278 \\\hline
        \noalign{\hrule height 1pt} 
	\end{tabular}
}
\end{center}
\end{table*}

\begin{figure*}[t]
\centering
\includegraphics[width=\columnwidth]{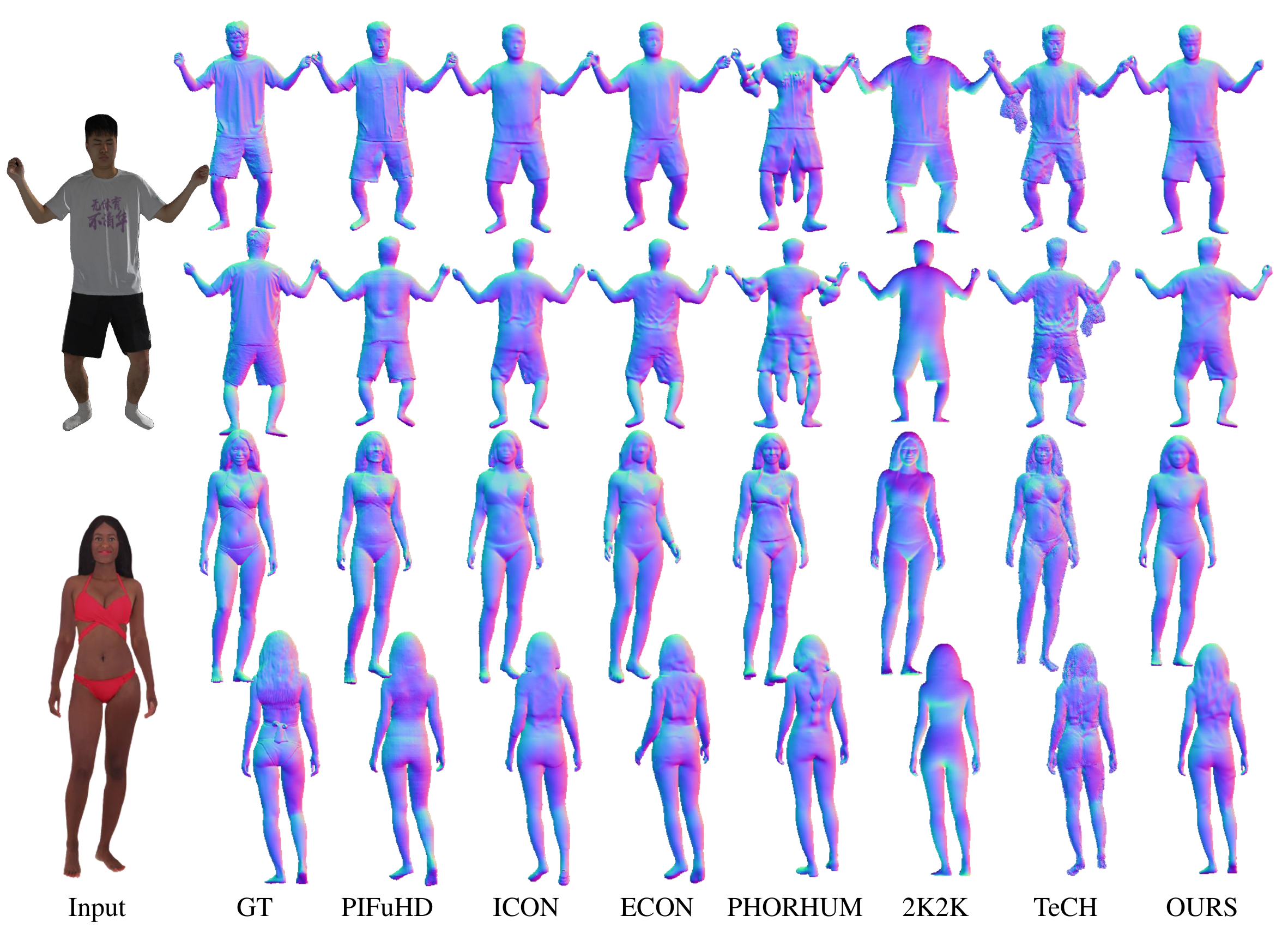}
\vspace{-8mm}
\caption{Comparison of normal maps. The first two rows are from the TH3.0 test data, and the latter two are from the RP dataset.}
\label{fig:posed_eval}
\end{figure*}

\subsection{Quantitative and Qualitative Evaluations}
We compare our method with existing static avatar reconstruction techniques, in \cref{table:eval_posed}. PHORHUM~\cite{alldieck2022photorealistic}, PaMIR~\cite{zheng2021pamir}, PIFu~\cite{saito2019pifu}, PIFuHD~\cite{saito2020pifuhd},  ICON~\cite{xiu2022icon}, 2K2K~\cite{han2023high}, ECON~\cite{xiu2023econ}, and TeCH~\cite{huang2023tech}. Here, PIFu$^{*}$, PaMIR$^{*}$, PIFuHD, and ICON were trained by using the RP dataset and ECON~\cite{xiu2023econ} used TH dataset, 2K2K~\cite{han2023high} used 2K2K and RP datasets, respectively. Since comparison methods are trained with different dataset settings, we train our model in several ways for a fair comparison: train only with RP, train only with TH, and train with both RP and TH. For methods that take SMPL guidance, we evaluate them with the ground truth SMPL-X that we used.

To assess human reconstruction methods, we employed average point-to-surface distance (P2S), Chamfer distance, and average surface normal error.
The units for P2S and Chamfer distances are centimeters. For normal errors, we projected normal maps to the front and back views, calculating the average L2 distance between ground truth and predicted normal maps. To enhance readability, the normal error values were multiplied by $100$. 

Our quantitative analysis presented in \cref{table:eval_posed} shows that our method performs better than previous methods. Our analysis indicates that leveraging diverse datasets consistently enhances performance, as evidenced by all three sets from our approach achieving the highest scores across all evaluation metrics. From this, we can argue that explicitly predicting depth maps from a single image with the guidance of SMPL-X remains an effective approach, without necessitating complex techniques.

\begin{table*}[t]
\setlength{\tabcolsep}{7pt}
\centering
\caption{Quantitative comparison with SCANimate~\cite{saito2021scanimate} in canonical space. We select 2 avatars from RP rigged T pose dataset and generate 15 posed scans for each avatar, where poses are obtained from CAPE dataset~\cite{CAPE:CVPR:20}. We differ the training sets to $5$ and $15$ views, and use 3D metrics for evaluation.}
\label{table:eval_canon}
\begin{center}
\footnotesize
\vspace{-7mm}
\resizebox{0.8\linewidth}{!}{%
	\begin{tabular}{l | c | c | c | c | c | c | c }
        \noalign{\hrule height 1pt}
        \multicolumn{2}{c|}{} & \multicolumn{3}{c|}{SET1} & \multicolumn{3}{c}{SET2} \\ \cline{3-8}
        \multicolumn{1}{c}{Methods} & \# of views & {P2S$\downarrow$} & {CF$\downarrow$} & {NR$\downarrow$} & {P2S$\downarrow$} & {CF$\downarrow$} & {NR$\downarrow$}
        \\\hline
        \multirow{2}{*}{SCANimate~\cite{saito2021scanimate}} & $5$ view
        & 1.362 & 1.316 & 2.921 & 1.076 & 1.039 & 2.071 \\
        & $15$ view
        & 1.103 & 1.124 & 2.453 & 0.997 & 0.971 & 1.939 \\
        \hline
        \multirow{2}{*}{Ours} & $5$ view
        & 0.244 & 0.260 & 0.482 & 0.180 & 0.209 & 0.395 \\
        & $15$ view
        & 0.199 & 0.218 & 0.412 & 0.149 & 0.175 & 0.307 \\
        \hline
        \noalign{\hrule height 1pt} 
	\end{tabular}
}
\end{center}
\vspace{-9mm}
\end{table*}

\begin{figure}[t]
\centering
\includegraphics[width=1.0\columnwidth]{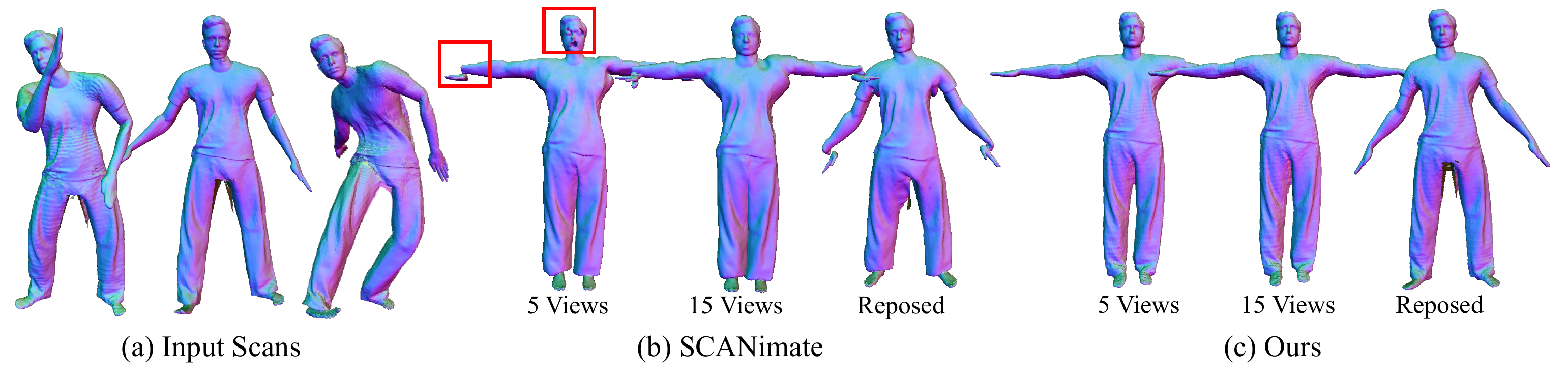}
\vspace{-8mm}
\caption{Comparison of results between SCANimate and our method. We used the same
SMPL pose parameters for SCANimate and ours. Five and fifteen scans were used to canonicalize the meshes.}
\label{fig:canon_eval}
\vspace{-2mm}
\end{figure}

\cref{fig:posed_eval} shows the qualitative results of ours and comparison results, where ECON, PHORHUM, 2K2K, and our methods recover the details in wrinkles and facial regions. The diversity and quantity of training data are crucial for model-free methods because they directly regress the human model from the image, in that PHORHUM frequently lost arm regions for the TH3.0 dataset, which contains diverse postures. In the case of 2K2K, they could not recover the details of the backsides because part-based normal prediction possibly loses the global context for the backside. TeCH utilizes normal map-based SDS loss~\cite{poole2022dreamfusion}, which often generates different facial shapes and undesirable artifacts on limbs.

We compare our framework with SCANimate~\cite{saito2021scanimate}, which takes a set of raw scans to generate a canonical mesh. SCANimate relies on the SMPL model to canonicalize the raw scans and to fill missing regions, therefore, it fails in the presence of pose errors. To compare SCANimate with our method, we took the same input as SCANimate, \ie, multiple raw scans, and used the same SMPL body pose parameters. The results are given in Tab.~\ref{table:eval_canon} and \cref{fig:canon_eval} that show our method accurately merges posed meshes into a canonical mesh.

\begin{figure*}[t]
\centering
\includegraphics[width=\columnwidth]{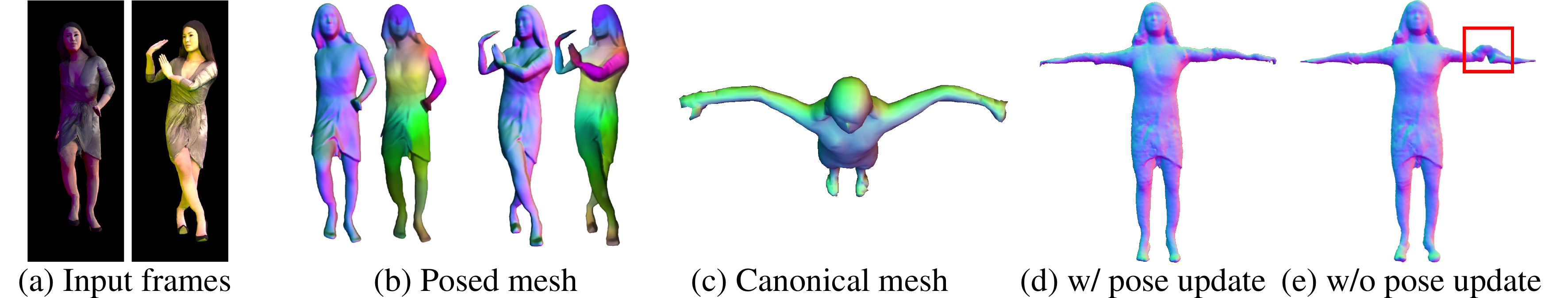}
\vspace{-7mm}
\caption{Reconstructed results with and without pose error refinement when the canonical mesh is given with noisy pose.}
\label{fig:abl_pose}
\end{figure*}

\begin{figure*}[t]
\centering
\includegraphics[width=\columnwidth]{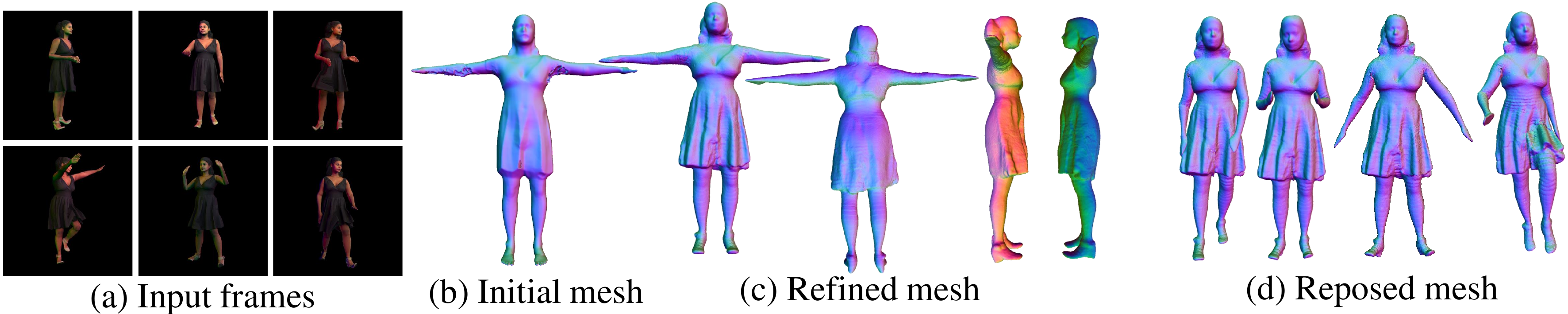}
\vspace{-7mm}
\caption{Example of reconstructed human wearing a loose cloth. 
(a) Given an input frame, (b) we initially reconstruct the canonical mesh. (c) With a forward skinning-based differentiable rendering, we integrate multiple frames and generate the refined avatar. (d) We can repose our mesh to unseen poses.}
\label{fig:loose_cloth}
\end{figure*}

\setlength\columnsep{5pt}
\begin{table}[t]
\begin{minipage}{0.5\columnwidth}
\includegraphics[width=\columnwidth]{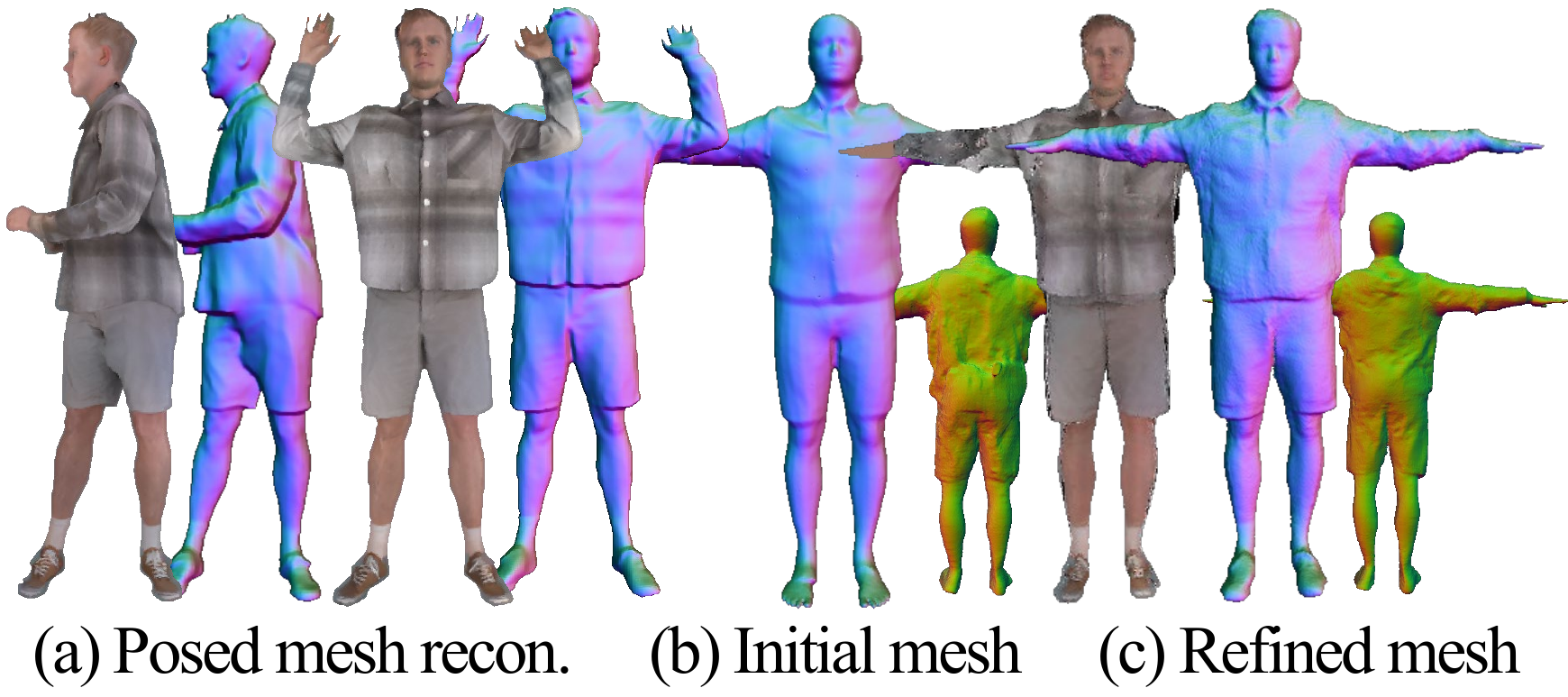}
\captionof{figure}{Actors-HQ results.}
\label{fig:actorhq}
\end{minipage}%
\hfill
\begin{minipage}{.5\columnwidth}
\includegraphics[width=1.\columnwidth]{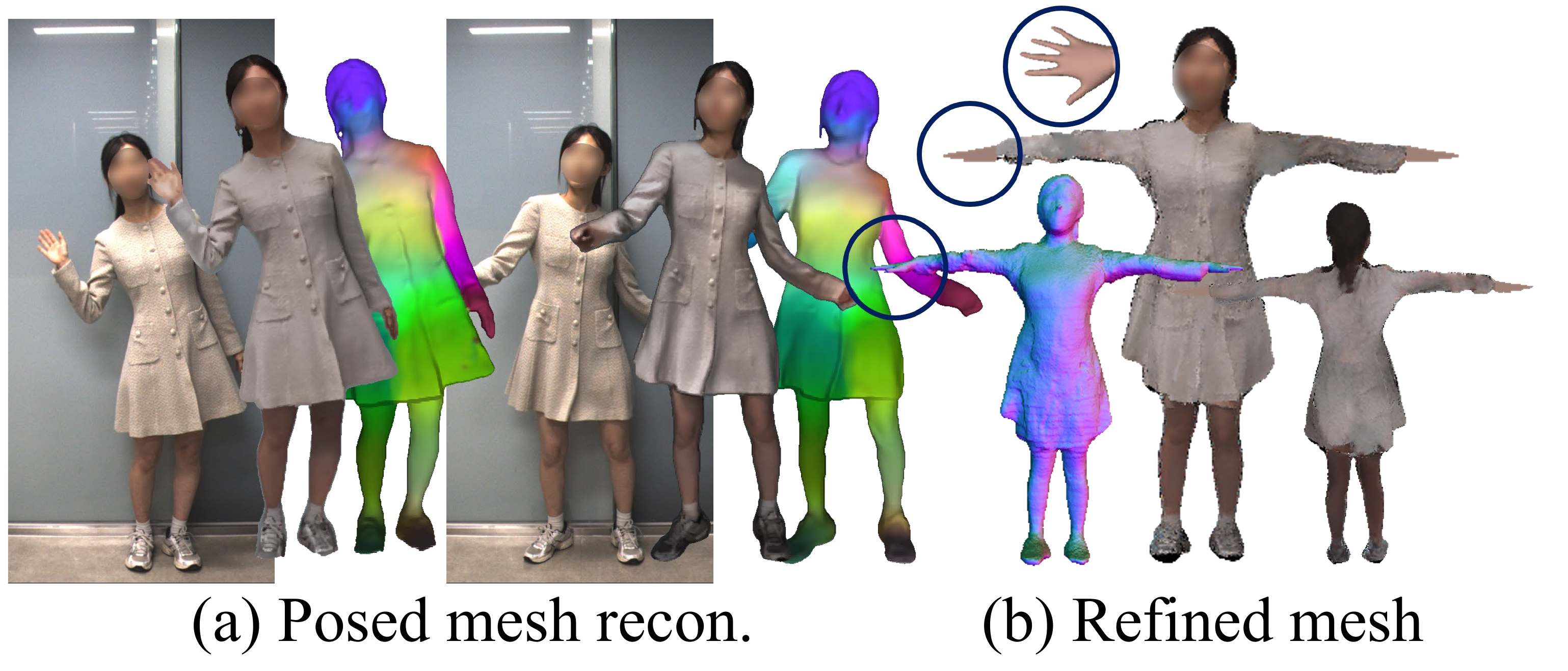}
\captionof{figure}{In-the-wild loose cloth results.}
\label{fig:inthewild}
\end{minipage}
\vspace{-15pt}
\end{table}

\subsection{Ablation Study}
We conduct several experiments to further analyze the characteristics of our proposed method. We demonstrate its robustness under pose variations and its ability to capture loose cloth, leveraging the advantage of the usage of the initial canonicalized model. 
In general, multiple frames improve the completeness of the human model because invisible regions shrink. Both \cref{fig:abl_pose} and \cref{fig:loose_cloth} indicate that our framework effectively integrates multiple results in the canonical space, even with errors in depth, LBS weights and joint positions.
We present results for challenging cases in \cref{fig:abl_pose}, where a woman shows difficult poses. In this case, the initial canonical mesh exhibits bent arms in \cref{fig:abl_pose}(e). Our differentiable rendering scheme, however, alleviates this error because it is able to refine pose errors as well as the shape and color. 
Furthermore, several studies~\cite{kim2023chupa, huang2023tech} employ differential rendering to obtain clothed human meshes, starting from a canonical template mesh. This means that it often fails to recover an accurate shape when the topology of the target mesh significantly differs from the template mesh. Contrarily, our initial canonical mesh is in a similar shape to the target mesh because we warp initial posed meshes into canonical space with predicted LBS weights, as shown in \cref{fig:loose_cloth}. 
Importantly, our method does not require ground-truth canonical meshes, as opposed to ARCH and ARCH++~\cite{huang2020arch, he2021arch++}. We only compare errors \wrt the images and normal maps in the posed space. \\
We also carry out experiments on real-world scenarios, including non-rigid cloth deformations in \cref{fig:actorhq} and \cref{fig:inthewild}, whose avatars are generated from Actors-HQ dataset~\cite{icsik2023humanrf} (Actor 8) and in-the-wild captured images, respectively. Here, we adopt hand replacement module from ECON~\cite{xiu2023econ} using the template model to demonstrate the capability of the hand refinement. Even with a large number of input frames, the blurry results occur when the cloth deformation is large, our approach can generate realistic 3D avatar with only a few frames. The experimental details and more results are provided in the supplementary material.

\section{Conclusion}
We have proposed CanonicalFusion which creates animatable human avatars from images via posed mesh prediction and forward skinning-based differentiable rendering. As a first step, we predict depth maps and LBS weights using a shared-encoder-dual-decoder network, where the LBS weights are compressed into a 3-dimensional latent vector without a loss of accuracy. Furthermore, we proposed a differentiable rendering technique to refine a canonical mesh by minimizing normal, color, and 3D human pose errors for an arbitrary number of images. We experimentally validated the effectiveness of our method from various aspects compared to existing methods. We believe our next direction is to handle the non-rigid deformation of cloth and hairs to simulate more realistic results and to combine our method with generative techniques to diversify our avatars.   

\noindent\textbf{Acknowledgements} This research was supported by the Institute of Information $\&$ communications Technology Planning $\&$ Evaluation (IITP) grant funded by the Korea government (MSIT) (No.2019-0-01842, Artificial Intelligence Graduate School Program (GIST) / No.2022-0-00566. The development of object media technology based on multiple video sources / RS-2024-00398830. The development of real-time digital human synthesis technology for real-world videos), the National Research Foundation of Korea (NRF) grant funded by the Korea government (MSIT)(RS-2024-00338439), the Ministry of Trade, Industry and Energy (MOTIE) and Korea Institute for Advancement of Technology (KIAT) through the International Cooperative R$\&$D program in part (P0019797), and the Korea Agency for Infrastructure Technology Advancement (KAIA) grant funded by the Ministry of Land Infrastructure and Transport (Grant RS-2023-00256888).


%
%
\bibliographystyle{splncs04}
\bibliography{main}
\end{document}